\documentclass{elsarticle}
\usepackage{graphicx}
\usepackage{amsmath}
\usepackage{amssymb}
\usepackage{latexsym}
\newtheorem{thm}{Theorem}
\newtheorem{problem}[thm]{Problem}
\newtheorem{defn}[thm]{Definition}
\newtheorem{exmp}[thm]{Example}

\usepackage{algorithm}
\usepackage{algorithmic}
\usepackage{subfigure}

\begin{document}

\begin{frontmatter}


\title{
Feature selection with test cost constraint
}
\author[fmin]{Fan Min\corref{cor1}}
\ead{minfanphd@163.com}
\author[qhu]{Qinghua Hu}
\ead{huqinghua@hit.edu.cn}
\author[fmin]{William Zhu}
\ead{williamfengzhu@gmail.com}

\cortext[cor1]{Corresponding author. Tel.: +86 133 7690 8359}

\address[fmin]{Lab of Granular Computing,
Zhangzhou Normal University, Zhangzhou 363000, China}
\address[qhu]{Tianjin University, Tianjin 300072, China}


\begin{abstract}
Feature selection is an important preprocessing step in machine learning and data mining.
In real-world applications, costs, including money, time and other resources, are required to acquire the features.
In some cases, there is a test cost constraint due to limited resources.
We shall deliberately select an informative and cheap feature subset for classification.
This paper proposes the feature selection with test cost constraint problem for this issue.
The new problem has a simple form while described as a constraint satisfaction problem (CSP).
Backtracking is a general algorithm for CSP, and it is efficient in solving the new problem on medium-sized data.
As the backtracking algorithm is not scalable to large datasets, a heuristic algorithm is also developed.
Experimental results show that the heuristic algorithm can find the optimal solution in most cases.
We also redefine some existing feature selection problems in rough sets, especially in decision-theoretic rough sets, from the viewpoint of CSP.
These new definitions provide insight to some new research directions.
\end{abstract}

\end{frontmatter}

  %
  %%%%%%%%%%%%%%%%%%%%%%%%%%%%%%%%%%
  % beginning of a new section: introduction
  %%%%%%%%%%%%%%%%%%%%%%%%%%%%%%%%%%
  %
  \section{Introduction}\label{section: introduction}
Many data mining approaches employ feature selection techniques to speed up learning and to improve model quality
\cite{HuQ2008MixedFeature,KiraK1992Feature,ZhongDongOhsuga01Using}.
These techniques are especially important for datasets with tens or hundreds of thousands of features \cite{GuyonI2003Introduction}.
Attribute reduction \cite{Pawlak82Rough} is a special type of feature selection problems studied by the rough set society.
A reduct is a feature subset that is jointly sufficient and individually necessary to preserve certain information of the data \cite{YaoZhao08Attribute}.
For decision making, the most often addressed information is the positive region with respect to the decision class \cite{Pawlak82Rough}.
The objective of the classical reduct problem is to find a minimal reduct \cite{SkowronRauszer92TheDiscernibility}, since simpler representation often provides better generalization ability according to Occam's razor principle.
Other feature selection problems aim at finding feature subsets with maximal margin \cite{CortesVapnik95Support}, maximal stability \cite{BazanSkowron94Dynamic}, minimal space \cite{MinDuQiuLiu07Minimal}, etc.

Most of these problems assume the data are already stored in datasets and available without charge.
However, data are not free in real-world applications.
There are test costs, such as money, time, or other resources \cite{MinHeQianZhu11Test,TurneyPD1995Cost} to obtain feature values of objects.
For example, it takes both time and money to obtain medical data of a patient \cite{ZhangSC2010Waiting}.
Under this context, one would like to select the cheapest reduct \cite{SusmagaR1999Computation}.
This consideration and the parallel test assumption have motivated the minimal test cost reduct (MTR)
problem \cite{MinHeQianZhu11Test}.
Recently, a number of algorithms have been developed to deal with this problem (see, e.g., \cite{HeMin11Accumulated,MinHeQianZhu11Test,PanMinZhu11Genetic}).
Other related issues have also been identified in addressing numerical features \cite{ZhaoMinZhu2011Test}, observational errors \cite{MinZhu12Tcsdser}, and test costs relationships \cite{HeMinZhu11Attribute,MinLiu09AHierarchical}.
All these problems aim at searching the cheapest feature subset which preserves sufficient information for classification.

Nevertheless, the available resource is usually limited, and users have to sacrifice necessary information to keep the test cost under budget.
This paper introduces the feature selection with test cost constraint (FSTC) problem to formulate this issue.
The upper bound of the available resource serves as the constraint.
The FSTC problem is more general than MTR \cite{MinHeQianZhu11Test}.
In fact, these two problems coincide when the constraint is no less than the test cost of the optimal reduct.
If the constraint is so tight that the sufficiency condition cannot be met, then one cannot obtain a reduct.
This is why the new problem falls in \emph{feature selection} instead of in \emph{attribute reduction}.

In this paper, the FSTC problem is defined from the viewpoint of the constraint satisfaction problem (CSP).
In other words, it is defined with four aspects, namely input, output, constraint, and optimization objective.
The new definition is simpler and easier to comprehend than the one defined from the viewpoint of set family \cite{MinZhu11Optimal}.
Furthermore, we redefine the classical reduct problem and the minimal reduct problem \cite{SkowronRauszer92TheDiscernibility} from the CSP viewpoint.
We show that most feature selection problems in rough sets, including those of decision-theoretic rough sets (DTRS) \cite{LiuLiLi12Multiple,YangYao12Modelling,YaoWong92ADecision,YaoZhao08Attribute,YaoZhaoWang08OnReduct}, can be viewed as extensions of the minimal reduct problem \cite{SkowronRauszer92TheDiscernibility} from one or more of these four aspects.
This viewpoint gives insight to meaningful research trends concerning feature selection in a broader sense.
In fact, there are some similar viewpoints, including the optimization viewpoint of attribute reduction on DTRS discussed by Jia et al. \cite{JiaX2011Optimization}.
Compared with them, the one presented here is more systematic.

We develop a backtracking algorithm to the FSTC problem for small and medium-sized datasets.
Backtracking algorithms are natural and effective approaches to CSPs for obtaining one or all optimal solutions.
However, they are seldom employed to deal with feature selection problems in rough set theory
(see, e.g., \cite{ChenYM2011Rough,MinZhu11OptimalDynamic,MinZhu12Tcsdser}), where discernibility matrix based approaches are more popular (see, e.g., \cite{QianMiaoZhangLi12Hybrid,SkowronRauszer92TheDiscernibility,WangWang01Reduction,YeChen02ANew}).
One possible reason is that people have not defined attribute reduction problems explicitly as CSPs.
As an exhaustive algorithm, the backtracking algorithm has a time complexity exponential with respect to the number of features.

We also develop a heuristic algorithm with polynomial time complexity for large datasets.
We employ the addition-deletion approach \cite{YaoZhaoWang08OnReduct} to design a heuristic function based
on information gain often employed in similar problems \cite{DaiXu2012Approximations,Slezak02Approximate,Wang02Attribute,YaoZhaoWang08OnReduct}.
It is similar to the one proposed in \cite{MinHeQianZhu11Test} to prefer low cost features through $\lambda$-weighting, where $\lambda$ is a user specified parameter.
The difference between the new algorithm and the one employed in \cite{MinHeQianZhu11Test} lies in the stopping criteria.
To improve the performance of the algorithm, we employ the competition strategy \cite{MinHeQianZhu11Test}.
With this strategy, different feature subsets are obtained through setting different $\lambda$ values, then the best one is selected.
This strategy can trade the quality of the result with the run time.
More importantly, with this strategy, the user is not involved in the setting of $\lambda$.
Instead, a set of $\lambda$ values which are valid for any dataset are specified by the algorithm.

Four open datasets are employed to study the performance of our algorithms.
Experimental results show that the backtracking algorithm is efficient for medium-sized data.
It takes less than 0.4 second to obtain an optimal feature subset for the mushroom dataset, which contains 22 features and 8124 objects.
The backtracking algorithm is approximately 10 times faster than SESRA \cite{MinZhu11Optimal}, which is based on another definition of the problem.
The heuristic algorithm is stably more efficient than the backtracking one.
With the help of the competition strategy, the heuristic algorithm can find the optimal solution in most cases.

The rest of the paper is organized as follows: Section \ref{section: problem-definition} presents the problem definition.
The classical reduct problem and the minimal test cost reduct problem are also redefined.
Section \ref{section: algorithms} proposes both backtracking and heuristic algorithms. Experimental results on four UCI (University of California - Irvine) datasets are discussed in Section \ref{section: experiments}.
Then Section \ref{section: discussions} studies existing feature selection problems in the rough set society from the viewpoint of CSP.
Some interesting new problems are also briefly discussed.
Finally, Section \ref{section: conclusion} presents the concluding remarks and further research directions.

  %
  %%%%%%%%%%%%%%%%%%%%%%%%%%%%%%%%%%
  % beginning of a new subsection
  %%%%%%%%%%%%%%%%%%%%%%%%%%%%%%%%%%
  %
\section{Problem definition}\label{section: problem-definition}
This section reviews three feature selection problems in rough sets.
Two of them are under the classical rough sets \cite{Pawlak82Rough},
and the last one is concerned with test cost
\cite{MinHeQianZhu11Test}. These problems are redefined as CSPs.
Moreover, we propose a new problem called feature selection with
test cost constraint.

  %
  %%%%%%%%%%%%%%%%%%%%%%%%%%%%%%%%%%
  % beginning of a new subsection
  %%%%%%%%%%%%%%%%%%%%%%%%%%%%%%%%%%
  %
\subsection{Classical feature selection problems in rough sets}
Data models are fundamental for feature selection. This paper only
considers decision systems.
\begin{defn}\label{defn: ds}\cite{Yao04APartition}
A \emph{decision system} (DS) $S$ is the 5-tuple:
\begin{equation}\label{equation: ds}
S = (U, C, d, V = \{V_a | a \in C \cup \{d\}\}, I = \{I_a | a \in C \cup \{d\}\}),
\end{equation}
where $U$ is a finite set of objects called the universe, $C$ is the set of features, $d$ is the decision class, $V_a$ is the set of values for each $a \in C \cup \{d\}$, and $I_a : U \rightarrow V_a$ is an information function for each $a \in C \cup \{d\}$.
\end{defn}

Let the decision system $S = (U, C, d, V, I)$ be nominal, that is, all features in $C$ are nominal.
Any $\emptyset \neq B \subseteq C \cup \{d\}$ determines an indiscernibility relation $I(B)$ on $U$.
A partition determined by $B$ is denoted by $U/I(B)$, or $U/B$ for brevity.
Let $\underline{B}(X)$ denote the $B-$\emph{lower approximation} of $X$.
The positive region of $\{d\}$ with respect to $B \subseteq C$ is defined as $POS_B(\{d\}) = \mathop \bigcup_{X \in
U/\{d\}}\underline{B}(X)$ \cite{Pawlak82Rough,Pawlak91Rough}.

\begin{defn}\label{defn: relative-reduct}\cite{Pawlak91Rough}
Any $B \subseteq C$ is called a \emph{decision relative reduct} (or
\emph{a reduct} for short) of $S$ iff:
\begin{enumerate}
\item{$POS_B(\{d\}) = POS_C(\{d\})$; and}
\item{$\forall a \in B, POS_{B - \{a\}}(\{d\}) \subset POS_C(\{d\})$}.
\end{enumerate}
\end{defn}

Definition \ref{defn: relative-reduct} indicates that a reduct is 1) jointly sufficient and 2) individually necessary for preserving a particular property (positive region in this context) of the decision system
\cite{LiHX11Further,Pawlak82Rough,YaoZhao08Attribute,ZhaoLuoWonYao07AGeneral}.
In other words, there are two constraint, named sufficiency and necessity, respectively.
Consequently, the problem of obtaining one reduct can be defined in the CSP style as follows.
\begin{problem}\label{problem: reduct}
The attribute reduction problem.\\
Input: $S = (U, C, d, V , I)$;\\
Output: $B \subseteq C$;\\
Constraints: (1) $POS_B(\{d\}) = POS_C(\{d\})$;\\(2) $\forall a \in B, POS_{B - \{a\}}(\{d\}) \subset POS_C(\{d\})$.\\
\end{problem}

There may exist many reducts for a decision system. Let the set of
all relative reducts of $S$ be $Red(S)$. Any $R \in Red(S)$ is a
minimal reduct if and only if $|R|$ is minimal. Minimal reducts are
preferred because they provide the simplest representation of the
knowledge. The problem of finding a minimal reduct is called the
minimal reduct problem, as defined as follows.
\begin{problem}\label{problem: minimal-reduct}
The minimal reduct problem.\\
Input: $S = (U, C, d, V , I)$;\\
Output: $B \subseteq C$;\\
Constraint: $POS_B(\{d\}) = POS_C(\{d\})$;\\
Optimization objective: $\min |B|$.
\end{problem}

Problem \ref{problem: minimal-reduct} has an optimization objective,
which is typical in CSP. Note that that there is only one
constraint, namely sufficiency. This does not indicate that the
necessity constraint is not met. In fact, the necessity constraint
can be derived from the optimization objective. One can easily prove
this by contradiction. That is, if there are superfluous features,
the size of the feature subset cannot be minimal. In other words,
the problem definition is simplified while viewed as a CSP.

  %
  %%%%%%%%%%%%%%%%%%%%%%%%%%%%%%%%%%
  % beginning of a new subsection
  %%%%%%%%%%%%%%%%%%%%%%%%%%%%%%%%%%
  %
\subsection{Feature selection minimizing test cost}
Test cost is an important issue in many applications. We have built
a hierarchy of six test-cost-sensitive decision systems
\cite{MinLiu09AHierarchical}. Here we present a simple model which
will be used in defining the new problem of this paper.
\begin{defn}\label{defn: cost-independent-ds}
\cite{MinLiu09AHierarchical} A \emph{test-cost-independent
decision system} (TCI-DS) $S$ is the 6-tuple:
\begin{equation}\label{equation: ci-ds}
S = (U, C, d, \{V_a | a \in C \cup \{d\}\}, \{I_a | a \in C \cup \{d\}\},
c),
\end{equation}
where $U, C, d, \{V_a\}$, and $\{I_a\}$ have the same meanings as in Definition \ref{defn: ds},
$c: C \rightarrow \mathbb{R}^+ \cup \{0\}$ is the test cost function.
Test costs are independent of one another, that is, $c(B) = \sum_{a \in B}c(a)$ for any $B \subseteq C$.
\end{defn}

The minimal test cost reduct (MTR) problem proposed in \cite{MinHeQianZhu11Test} can be redefined as follows.
\begin{problem}\label{problem: minimal-test-cost-reduct}
The minimal reduct problem.\\
Input: $S = (U, C, d, V , I, c)$;\\
Output: $B \subseteq C$;\\
Constraint: $POS_B(\{d\}) = POS_C(\{d\})$;\\
Optimization objective: $\min |c(B)|$.
\end{problem}

One can see there are two differences between Problem \ref{problem:
minimal-test-cost-reduct} and Problem \ref{problem: minimal-reduct}.
The first is the input, where the test cost is the external
information. The second is the optimization objective, which is to
minimize the test cost, instead of the number of features.

  %
  %%%%%%%%%%%%%%%%%%%%%%%%%%%%%%%%%%
  % beginning of a new subsection
  %%%%%%%%%%%%%%%%%%%%%%%%%%%%%%%%%%
  %
\subsection{Feature selection with test cost constraint}
Sometimes we are given limited resources to obtain the feature values.
We proposed the issue of optimal sub-reduct in \cite{MinZhu11Attribute,MinZhu11Optimal} to address this issue.
Here we use the positive region instead of the conditional information entropy to define respective concepts.
\begin{defn}\label{defn: optimal-sub-reduct}
Let $S = (U, C, d, V, I, c)$ be a TCI-DS and $m$ the test cost upper bound.
The set of all feature subsets subject to the constraint is
\begin{equation}\label{equation: test-cost-constraint-sets}
T(S, m) = \{B \subseteq C | c(B) \leq m\}.
\end{equation}
In $T(S, m)$, the set of all feature subsets with the maximal positive region is
\begin{equation}\label{equation: maximal-pos-rate}
M_T(S, m) = \{B \in T(S, m) | POS_B(\{d\}) = \min\{POS_{B'}(\{d\}) | B' \in T(S, m)\}\}.
\end{equation}
In $M_T(S, m)$, the set of all optimal sub-reducts is
\begin{equation}\label{equation: least-cost-sets}
P_{M_T}(S, m) = \{B \in M_T(S, m) | c(B) = \min\{c(B') | B' \in M_T(S, m)\}\}.
\end{equation}
Any element in $P_{M_T}(S, m)$ is called an optimal sub-reduct with test cost constraint, or an optimal sub-reduct for brevity.
\end{defn}

In Definition \ref{defn: optimal-sub-reduct}, Equation (\ref{equation: test-cost-constraint-sets}) ensures the constraint is met;
Equation (\ref{equation: maximal-pos-rate}) ensures most informative feature subset is selected;
and Equation (\ref{equation: least-cost-sets}) ensures test cost is minimized. The problem of
constructing $P_{M_T}(S, m)$ is called the \emph{optimal sub-reducts
with test cost constraint} (OSRT) problem
\cite{MinZhu11Attribute,MinZhu11Optimal}. Unfortunately, the
definition is rather prolonged and hard to read. Next we follow the
style of Problem \ref{problem: minimal-reduct} to present the
following problem.
\begin{problem}\label{problem: test-cost-constraint-reduct}
The feature selection with test cost constraint (FSTC) problem.\\
Input: $S = (U, C, d, V , I, c)$, the test cost upper bound $m$;\\
Output: $B \subseteq C$;\\
Constraint: $c(B) \leq m$;\\
Optimization objectives: (1) $\max POS_B(\{d\})$; and (2) $\min c(B)$.
\end{problem}

Note that the two objectives are not equally important.
They are the primary and the secondary objectives, respectively.
In fact, Problem \ref{problem: test-cost-constraint-reduct} is the same as the OSRT problem.
However the problem definition is simpler and easier to comprehend.
This phenomenon indicates that the form of CSP is more appropriate for this kind of problems.

By comparing Problems \ref{problem: minimal-test-cost-reduct} and \ref{problem: test-cost-constraint-reduct}, we observe the following differences.
First, the constraint is expressed by the test cost instead of the positive region.
Second, the first objective of Problem \ref{problem: test-cost-constraint-reduct} is to maximize the positive region. Third, the objective of Problem \ref{problem: minimal-test-cost-reduct} becomes the secondary objective of Problem
\ref{problem: test-cost-constraint-reduct}.
This objective is considered after the primary one is achieved.

In fact, Problem \ref{problem: test-cost-constraint-reduct} is more general than Problem \ref{problem: minimal-test-cost-reduct}.
Let $B'$ be a minimal test cost reduct subject to Problem \ref{problem: minimal-test-cost-reduct}.
If $m \geq c(B')$, the constraint is met when the primary objective is achieved.
In other words, the constraint is essentially redundant.
The first objective will be replaced by $POS_B(\{d\}) = POS_C(\{d\})$, which serves as a constraint.
The second objective is then the only objective.
Consequently, Problem \ref{problem: test-cost-constraint-reduct} coincides with Problem \ref{problem: minimal-test-cost-reduct} in this case.

  %
  %%%%%%%%%%%%%%%%%%%%%%%%%%%%%%%%%%
  % beginning of a new section: conclusions
  %%%%%%%%%%%%%%%%%%%%%%%%%%%%%%%%%%
  %
  \section{Algorithm design}\label{section: algorithms}
This section presents two algorithms. One is a backtracking algorithm,
the other is a heuristic algorithm. The backtracking algorithm always
produces an optimal solution to the problem. The heuristic algorithm
is more efficient to large datasets, however the feature subset
obtained may not be optimal.

\subsection{The backtracking algorithm}\label{subsection: back-track}
The backtracking algorithm is a natural solution to CSP.
In the rough set society, people seldom employ this algorithm for attribute reduction.
This is partly due to the form of problem definition as shown in Definition \ref{defn: relative-reduct}.
The backtracking algorithm to the FSTC problem is illustrated in Algorithm \ref{algorithm: backtracking}.
To invoke the algorithm, one should initialize the global variables $m$, let $B = \emptyset$, and use the following statement:\\
backtracking($\emptyset$, 0);\\
then at the end of the algorithm execution, an optimal feature subset will be stored in $B$.

\begin{algorithm}[tb!]\caption{The backtracking algorithm to the FSTC problem}\label{algorithm: backtracking}
  \textbf{Input}: Selected feature subset $B'$, feature index lower bound $l$\\
  \textbf{Output}: Results are stored in the global variable $B$\\
  \textbf{Method}: backtracking
  \begin{algorithmic}[1]
        \FOR {($i = l$; $i < |C|$; $i$ ++)}
            \STATE $B'' = B' \cup \{a_i\}$;//One more feature
            \IF {($c(B'') > m$)}
                \STATE continue;//The constraint is violated
            \ENDIF

            \IF {($POS_{B''}(\{d\}) = POS_C(\{d\})$)}
                \STATE throw new Exception(``Coincides with the MTR problem");
            \ENDIF

            \IF {($|POS_{B''}(\{d\})| > |POS_B(\{d\})|$ $\vee$ $(POS_{B''}(\{d\}) = POS_B(\{d\}))$ $\wedge$ $(c(B'') < c(B))$)}
                \STATE $B = B''$;//A better feature subset
            \ENDIF

            \STATE backtracking($B''$, $i + 1$);//Backtracking
       \ENDFOR
  \end{algorithmic}
\end{algorithm}

In Algorithm 1, Lines 3 through 5 check the constraint.
Feature subsets violating the constraint are simply discarded.
Lines 6 through 8 indicate if the positive region of the current feature subset is the same as $C$, namely the sufficiency condition can be met, the FSTC problem coincides with the MTR problem.
In this case we only need to address the MTR problem.
Lines 9 through 11 are devoted to the optimization objective.
$|POS_{B''}(\{d\})| > |POS_B(\{d\})|$ serves for the first objective.
$c(B'') < c(B)$ serves for the second; it is checked only if $POS_{B''}(\{d\}) = POS_B(\{d\})$.
In our implementation in Coser \cite{Coser}, the algorithm is implemented to avoid repeated computation of positive regions.

Note that a feature is never removed from a subset.
This is important to ensure the correctness of the algorithm.
Line 2 shows that feature $a_i$ is added.
It may happens that $POS_{B''}(\{d\}) = POS_{B'' \cup \{a_i\}}(\{d\})$, i.e., $a_i$ does not contribute to the positive region.
However, $a_i$ is not removed because it may be useful while combined with other features.
We introduce the following example to explain the reason.
\begin{exmp}\label{example: pos-inf-invalid}
Consider the decision system listed in Table \ref{table: pos-inf-invalid}.
Let $c = [2, 3, 10]$ and $m = 6$.
Because $c(a_3) = 10 > m$, $a_3$ is never selected.
We have\newline $POS_{\{a_1\}}(\{d\}) = POS_{\{a_2\}}(\{d\}) = \emptyset$.
That is, neither $a_1$ nor $a_2$ contributes to the positive region alone.
However, $POS_{\{a_1, a_2\}}(\{d\}) = \{x_2, x_3, x_4\}$, hence both $a_1$ and $a_2$ are useful.
The optimal feature subset is $\{a_1, a_2\}$, which is the output of the algorithm.
\end{exmp}

\setlength{\tabcolsep}{18pt}
\begin{table}[tb]\caption{A decision table for Example \ref{example: pos-inf-invalid}}
\label{table: pos-inf-invalid}
\begin{center}
\begin{tabular}{cccccccc}
\hline
$U$       & $a_1$   & $a_2$ & $a_3$ & $d$ \\
\hline
$x_1$     & Y       & Y     & Y     & A\\
$x_2$     & N       & Y     & N     & B\\
$x_3$     & Y       & N     & N     & B\\
$x_4$     & N       & N     & Y     & A\\
$x_5$     & Y       & Y     & Y     & B\\
\hline
\end{tabular}
\end{center}
\end{table}

In fact, $B$ may contain some redundant features during the algorithm execution.
It will eventually replaced by another feature subset with bigger positive region or smaller test cost in Line 10.
Example \ref{example: pos-inf-invalid} will be discussed further in Section \ref{subsection: heuristic}.

The space complexity of Algorithm \ref{algorithm: backtracking} is easy to analyze.
The algorithm searches in a tree with depth $|C|$ in a depth-first manner.
Whenever the backtracking method is invoked there is a need to obtain a new partition of the objects, which takes $O(|U| \times |C|)$ space.
Hence the space complexity is
\begin{equation}\label{equation: worst-space-backtracking}
O(|C| \times |U| \times |C|) = O(|U| \times |C|^2).
\end{equation}

Now we analyze the time complexity.
The number of feature subsets is $2^{|C|}$.
In the worst case all of them are checked.
On the other hand, a feature subset is never checked twice.
Therefore the number of backtracking steps, namely the number of time the backtracking method is invoked, is bounded by $2^{|C|}$.
As indicated by Line 1, each time we need to compute a feature subset with one more feature.
In this way, the computation involves splitting the dataset according to the current feature.
Respective operation takes $O(|U| \times |V_{a_i}|)$ of time.
Let $v_{max} = \max_{a \in C}|V_a|$.
The time complexity is
\begin{equation}\label{equation: worst-time-backtracking}
O(|U| \times 2^{|C|}\times v_{max}).
\end{equation}
Unfortunately, the average time complexity is hard to analyze.
We will show by experimentation that it is significantly lower than the worst case.

The design of the algorithm is often closely related to the problem definition.
Algorithm \ref{algorithm: backtracking} can be easily obtained from Problem \ref{problem: test-cost-constraint-reduct}.
Similarly, the SESRA algorithm \cite{MinZhu11Optimal} has three main steps, as indicated by Definition \ref{defn: optimal-sub-reduct}.
This phenomenon shows further the influence of the problem viewpoint to the problem definition and the algorithm design.

\subsection{The heuristic algorithm}\label{subsection: heuristic}
The backtracking algorithm is not scalable.
As indicated by Equation (\ref{equation: worst-time-backtracking}), the run time can be exponential with respect to the number of features in the worst case.
Hence we need to design heuristic algorithms for large datasets.
We adopt the well known addition-deletion approach \cite{MinLiu09AHierarchical,YaoZhaoWang08OnReduct} to design our algorithm, since the deletion approach is inefficient for large datasets \cite{YaoZhaoWang08OnReduct}.

The positive region seems to be a natural heuristic information, however, it may not work on some datasets.
Let $B$ be the currently selected feature subset.
We would like to select $a_i \in C - B$ if it is informative (i.e., $|POS_{B \cup \{a_i\}} - POS_{B}|$ is big) and cheap (i.e., $c(a_i)$ is small).
Unfortunately, we have counterexamples to this approach.
Let us consider Example \ref{example: pos-inf-invalid} again.
At the very beginning $B = \emptyset$.
Since $POS_{B \cup \{a_1\}} = \emptyset$, $a_1$ has no contribution to the positive region and therefore cannot be selected.
For the same reason $a_2$ is not selected.
$a_3$ cannot be selected due to the test cost constraint.
Finally, this approach fails to construct the optimal feature subset $\{a_1, a_2\}$.
Such cases happen in applications frequently.
We have tested this approach on four datasets listed in \ref{table: datasets}.
In the Voting and Tic-tac-toe datasets \cite{UCI}, no feature alone produces positive region, therefore the approach fails given any test cost setting.

A feasible heuristic information is the information gain \cite{Quinlan86Induction,WangYuYang02Decision}.
Generally, a feature subset with less information entropy tends to produce bigger positive region.
Therefore we employ information gain in this paper to design our algorithm.
Let $H(Q | P)$ be the conditional information entropy of $Q$ w.r.t. $P$
\cite{WangYuYang02Decision}. Let further $B \subset C$ and $a_i \in
C - B$, the information gain of $a_i$ w.r.t. $B$ is
\begin{equation}\label{equation: infomration-gain}
f_e(B, a_i) = H(\{d\} | B) - H(\{d\} | B \cup \{a_i\}).
\end{equation}
It is proven that $|POS_{B \cup \{a_i\}} - POS_{B}| > 0$ gives $H(\{d\} | B) - H(\{d\} | B \cup \{a_i\}) > 0$.
But the reverse does not hold.
In other words, information entropy is more sensitive to feature than positive region.

To select the current best feature, both information gain \cite{WangYuYang02Decision} and test cost are taken
into consideration.
We use the same approach as that in \cite{MinHeQianZhu11Test} to select the current best test.
And the $\lambda$-weighted function is defined as
\begin{equation}\label{equation: lambda-weighting}
f(B, a_i, c) = f_e(B, a_i)c_i^\lambda,
\end{equation}
where $\lambda$ is a non-positive number.
With the introduction of $\lambda$, cheaper features are preferred.
If $\lambda = 0$, $f(B, a_i, c) = f_e(B, a_i)$, and the heuristic information coincides with the information gain.

\begin{algorithm}[tb!]\caption{The $\lambda$-weighted heuristic algorithm}\label{algorithm: fstc-heuristic}
  \textbf{Input}: $S = (U, C, D, V, I, c)$, $m$\\
  \textbf{Output}: $B \subseteq C$\\
  \textbf{Method}: $\lambda$-weighted-fstc\\
  \begin{algorithmic}[1]
    \STATE $B = \emptyset$; //initialize the output
    \STATE $CA = C$; //unprocessed features
    \STATE $c_l = m$; //available test cost

    //Compute a feature subset with the least information entropy
    \WHILE {($CA \neq \emptyset$)}
      \STATE For any $a \in CA$ satisfying $C(a) \leq c_l$, compute $f(B, a, c)$;

      //Addition
      \STATE Select $a' \in CA$ with the maximal $f(B, a', c)$;
      \STATE $B = B \cup \{a'\}$; $CA = CA -\{a'\}$; $c_l = c_l - c(a')$;

      //Deletion, remove redundant features from the viewpoint of information entropy
      \FOR {(each $a \in B$)}
        \IF {($H({\{d\} | B - \{a\}}) = H(\{d\} | B)$)}
          \STATE $B = B - \{a'\}$; //$a'$ is redundant
          \STATE $c_l = c_l + c(a')$; //restore the constraint
        \ENDIF
      \ENDFOR

      //Remove features not satisfying the constraint to speed up
      \FOR {(each $a \in CA$)}
        \IF {($c_a > c_l$)}
          \STATE $CA = CA - \{a\}$;
        \ENDIF
      \ENDFOR
    \ENDWHILE

    //Remove redundant features from the viewpoint of positive region
      \FOR {(each $a \in B$)}
         \IF {($POS_{B - \{a'\}}(\{d\}) = POS_B(\{d\})$)}
            \STATE $B = B - \{a'\}$; //$a'$ is redundant
         \ENDIF
      \ENDFOR
    \STATE return $B$;
\end{algorithmic}
\end{algorithm}

Our algorithm is listed in Algorithm \ref{algorithm: fstc-heuristic}.
The algorithm first constructs a feature subset meeting the constraint and with minimal information entropy in Lines 4 through 19.
Lines 14 through 18 are not necessary, however they help speeding up the algorithm.
Then redundant features are removed from the viewpoint of the positive region in Lines 20 through 24.

One may find that the algorithm is successful on Example \ref{example: pos-inf-invalid}.
If we remove $x_5$ from the dataset, this algorithm also fails.
To make the matter worse, the ID3 decision tree encounters the same problem.
This might be a drawback of heuristic algorithms compared with exhaustive ones.
Fortunately, this extreme case seldom happens in applications.
On many UCI datasets we tested, Algorithm \ref{algorithm: fstc-heuristic} never fails to construct a feature subset.

The space complexity of Algorithm \ref{algorithm: fstc-heuristic} is decided by the size of the decision system.
It is
\begin{equation}\label{equation: space-algorithm-2}
O(|U| \times |C|).
\end{equation}
Now we analyze the time complexity.
In the worst case, the \textbf{while} loop indicated by Line 4 would execute $|C|$ times, and each time all remaining features are checked in Line 5.
Line 5 is executed at most $\sum_{i = 0}^{|C| - 1}(|C| - i) = O(|C|^2)$ times.
Since $f(B, a, c)$ is based on the positive region, similar to the analysis in Section \ref{subsection: back-track}, the time complexity is
\begin{equation}\label{equation: time-algorithm-2}
O(|U| \times |C|^2 \times v_{max}).
\end{equation}

In applications, it is hard for the user or even the expert to set a
rational $\lambda$. To make the matter worse, the best $\lambda$
does not always produce the best result. We can adopt the
competition strategy working as follows. First, it specifies a set
of $\lambda$ values, then it obtains corresponding feature subsets
using Algorithm \ref{algorithm: fstc-heuristic}, finally it chooses
the feature subset with the maximal positive region and the minimal
test cost. Since feature subsets produced by different $\lambda$
values compete against each other with only one winner, this
strategy is called the \emph{competition strategy}
\cite{MinHeQianZhu11Test}.

Formally, let $B_\lambda$ be the feature subset constructed by Algorithm \ref{algorithm: fstc-heuristic} using the exponential $\lambda$.
With $\Lambda$ the set of user-specified $\lambda$ values,
\begin{equation}\label{equation: best-among-lambda}
POS_\Lambda = \max_{\lambda \in \Lambda} POS_{B_\lambda}(\{d\})
\end{equation}
is the maximal positive region that can be obtained with the competition strategy.
This process requires the algorithm to be run $|\Lambda|$ times and the time complexity would be $O(|\Lambda| \times |U| \times |C|^2 \times v_{max})$ instead.
It is acceptable for relatively small $|\Lambda|$.
We will show that setting $\Lambda$ is easy in Section \ref{subsection: heuristic-effectiveness}.

  %
  %%%%%%%%%%%%%%%%%%%%%%%%%%%%%%%%%%
  % beginning of a new section: experiments
  %%%%%%%%%%%%%%%%%%%%%%%%%%%%%%%%%%
  %
  \section{Experiments}\label{section: experiments}
The main purpose of our experiments is to answer the following questions.
\begin{enumerate}
\item{Is the backtracking algorithm efficient?}
\item{Is the heuristic algorithm effective?}
\item{Is there an optimal setting of $\lambda$ for any dataset?}
\item{Is the extra computation time consumed by the competition strategy worthwhile?}
\end{enumerate}

  %
  %%%%%%%%%%%%%%%%%%%%%%%%%%%%%%%%%%
  % beginning of a new subsection: datasets
  %%%%%%%%%%%%%%%%%%%%%%%%%%%%%%%%%%
  %
  \subsection{Datasets}\label{subsection: datasets}
We deliberately select four datasets from the UCI Repository of Machine Learning Databases \cite{UCI}.
Their basic information is listed in Table \ref{table: datasets}, where $|C|$ is the number of features, $|U|$ is the number of instances, and $d$ is the name of the decision.

\setlength{\tabcolsep}{10pt}
\begin{table}[ht]\caption{Dataset information}
\label{table: datasets}
\begin{center}
\begin{tabular}{lrrrrr}
\hline
Name    & Domain   & $|C|$  & $|U|$   & $d$\\
\hline
Zoo     & zoology  & 16     & 101   & type\\
Voting  & society  & 16     & 435   & vote\\
Tic-tac-toe & game & 9      & 958   & class\\
Mushroom & botany  & 22     & 8124  & classes\\
\hline
\end{tabular}
\end{center}
\end{table}

There are a number of notes to make.
While counting the number of features, the decision is not included.
Missing values (e.g., those appearing in the Voting dataset) are treated as one particular value. That is, ? is equal to itself, and unequal to any other value.
The ``animal name" feature is not useful in the Zoo dataset, and we simply remove it.

Most datasets from the UCI library \cite{UCI} do not provide test cost information.
For statistical purposes, we need to produce them.
Different test cost distributions correspond to different applications.
Three distributions, namely uniform distribution, normal distribution, and Pareto distribution have been discussed in \cite{MinHeQianZhu11Test}.
For simplicity, this paper only employs the uniform distribution to generate random test cost in [1..100].
According to Definition \ref{defn: cost-independent-ds}, two TCI-DS are different once their test cost settings are different.
In this sense, we can produce as many TCI-DS as needed from a given DS.

  %
  %%%%%%%%%%%%%%%%%%%%%%%%%%%%%%%%%%
  % beginning of a new subsection
  %%%%%%%%%%%%%%%%%%%%%%%%%%%%%%%%%%
  %
  \subsection{Efficiency of the algorithms}\label{subsection: backtracking-efficiency}
We need to know the efficiency of the backtracking algorithm from three viewpoints.
The first is the average time complexity.
We need to know whether or not the number of backtracking steps is exponential with respect to the number of features.
The second is the time taken for small or medium-sized data.
In fact, diagnosis data for one particular disease in a hospital may contain only a few thousands of instances.
For those datasets, an optimal solution is always required.
The third is the run time compared with other exhaustive approaches.
The backtracking algorithm is compared with SESRA and SESRA$^*$ proposed in \cite{MinZhu11Optimal}.
SESRA is based on Definition \ref{defn: optimal-sub-reduct}, and SESRA$^*$ is an enhanced version.

\setlength{\tabcolsep}{5pt}
\begin{table}[tb]\caption{Backtracking steps on four datasets (with 100 test cost settings)}
\label{table: backtracking-steps}
\begin{center}
\begin{tabular}{lrrrrrrrr}
\hline
Dataset     &  $|C|$ & $2^{|C|}$  & \multicolumn{3}{c}{$|B|$}    &  \multicolumn{3}{c}{backtracking steps}\\
\cline{4-6} \cline{7-9}
            &        &            & min & max & av.  & min   & max    & av.\\
\hline
Zoo         &  16    & 65,536     & 4   & 6   & 4.74 & 132   & 4,089  & 1,112\\
Voting      &  16    & 65,536     & 7   & 9   & 8.23 & 8,139 & 46,421 & 24,354\\
Tic-tac-toe &  9     & 512        & 6   & 7   & 6.70  & 271   & 439   & 386\\
Mushroom    &  22    & 4,194,304  & 3   & 6   & 4.31 & 26    & 4,899  & 725\\
\hline
\end{tabular}
\end{center}
\end{table}

\setlength{\tabcolsep}{10pt}
\begin{table}[tb]\caption{Run time (ms) on four datasets (mean values for 100 test cost settings)}
\label{table: time-heuristic}
\begin{center}
\begin{tabular}{lrrrr}
\hline
Dataset     &  SESRA      & SESRA$^*$ & backtracking & heuristic\\
\hline
Zoo         &  50         & 48        & 7         & 2\\
Voting      &  5,334      & 2,498     & 485       & 18\\
Tic-tac-toe &  167        & 39        & 28        & 26\\
Mushroom    &  3,661      & 857       & 367       & 180\\
\hline
\end{tabular}
\end{center}
\end{table}

Table \ref{table: backtracking-steps} shows the number of backtracking steps, namely how many times the backtracking method is invoked.
Let $BS$ denote this number.
$2^{|C|}$ is the size of the backtracking tree, hence it is also the upper bound of $BS$.
For the Voting dataset, $|C| = 16$ and sometimes $|B| = 9$.
Therefore the maximal $BS$ can be 46,421, which is close to $2^{|C|} = 65,536$.
This indicates that sometimes $BS$ can be exponential with respect to $|C|$.
In contrast, For the Mushroom dataset, $|C| = 22$ and sometimes $|B| = 6$.
The maximal $BS$ is only 4,899, which is significantly smaller than $2^{|C|} = 4,194,304$.
In one word, $BS$ is relevant to not only $|C|$, but also $|B|$.

Table \ref{table: time-heuristic} compares the performance of the backtracking algorithm with SESRA and SESRA$^*$ \cite{MinZhu11Optimal} in terms of the run time.
The backtracking algorithm only takes 367 ms and 485 ms on the Mushroom and Voting datasets, respectively.
In other words, it is appropriate for many real applications.
Moreover, the backtracking algorithm stably outperforms SESRA and SESRA$^*$.
Only about 1/10 time is taken on the Tic-tac-toe and Mushroom datasets compared with SESRA.
These results show further the advantage of the CSP viewpoint.

For convenience, the run time of the heuristic algorithm is also listed in Table \ref{table: time-heuristic}.
The heuristic algorithm is always more efficient than exhaustive algorithms.
The efficiency difference becomes significant when the run time of exhaustive algorithms is long.
Moreover, the efficiency depends more on the dataset size instead of $|B|$.
To sum up, the heuristic algorithm can deal with larger datasets compared with exhaustive algorithms.

  %
  %%%%%%%%%%%%%%%%%%%%%%%%%%%%%%%%%%
  % beginning of a new subsection
  %%%%%%%%%%%%%%%%%%%%%%%%%%%%%%%%%%
  %
  \subsection{Effectiveness of the heuristic algorithm}\label{subsection: heuristic-effectiveness}
We compare the performance of the three approaches mentioned in Section \ref{subsection: heuristic}.
All three are based on Algorithm \ref{algorithm: fstc-heuristic}.
The first approach, called the non-weighting approach, is implemented by setting $\lambda = 0$.
The second approach, called the best $\lambda$ approach, chooses the best $\lambda$ value in $\Lambda$ = \{0, -0.25, -0.5, \dots, -3\}.
The third approach is the competition strategy based $\Lambda$ as discussed in Section \ref{subsection: heuristic}.

\begin{figure}[ht]
    \begin{center}
    \includegraphics[width=4.5in]{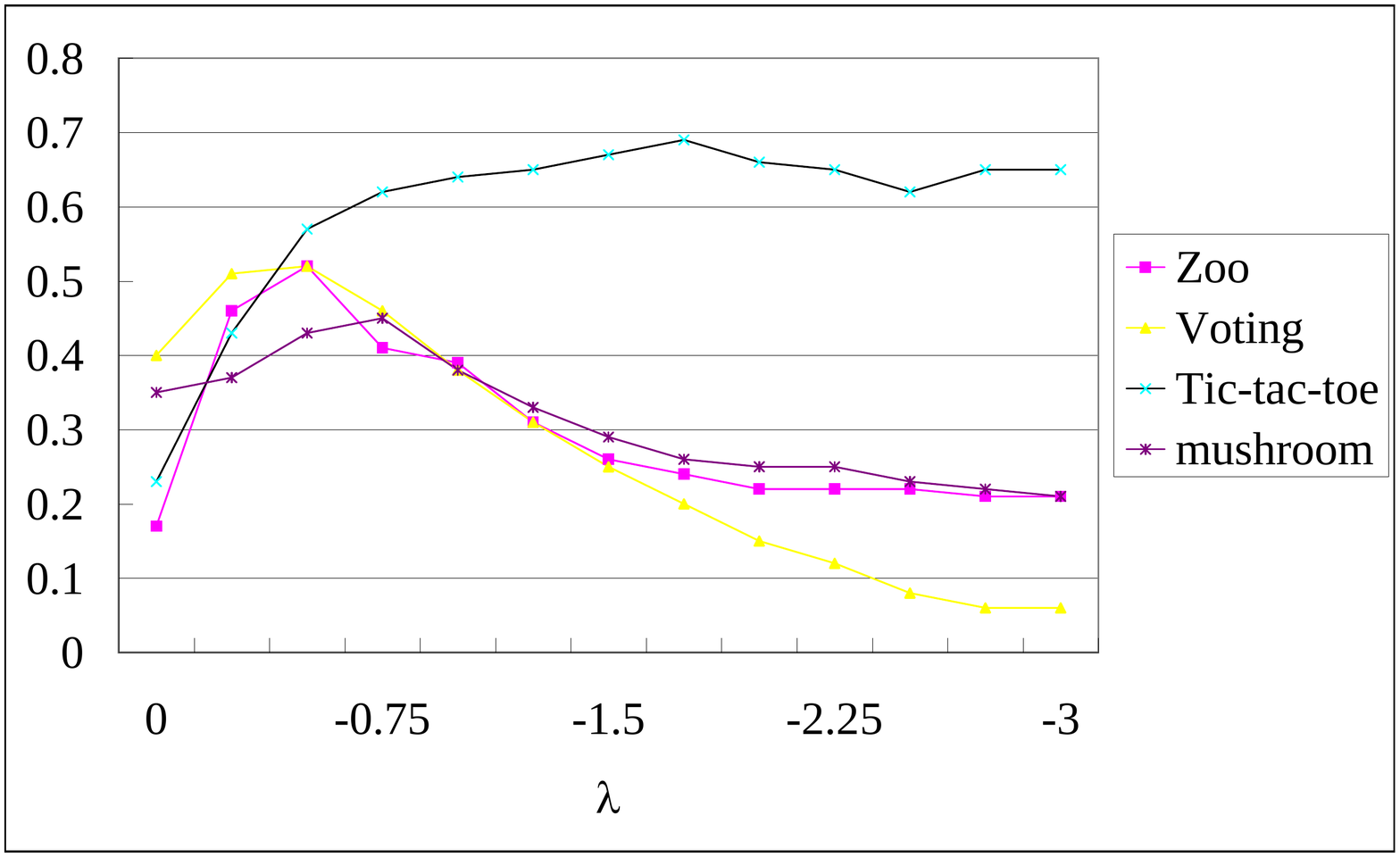}
    \caption{The probability of finding the optimal feature subset for given $\lambda$}
    \label{figure: lambda}
    \end{center}
\end{figure}

We now look at the influence of the $\lambda$ setting.
Fig. \ref{figure: lambda} shows the probability of finding the optimal feature subset for given $\lambda$.
Although $-0.75$ seems a reasonable value, there does not exist an optimal setting of $\lambda$ for all datasets.
In other words, $\lambda$ is hard to specify.

\begin{figure}[ht]
    \begin{center}
    \includegraphics[width=4.5in]{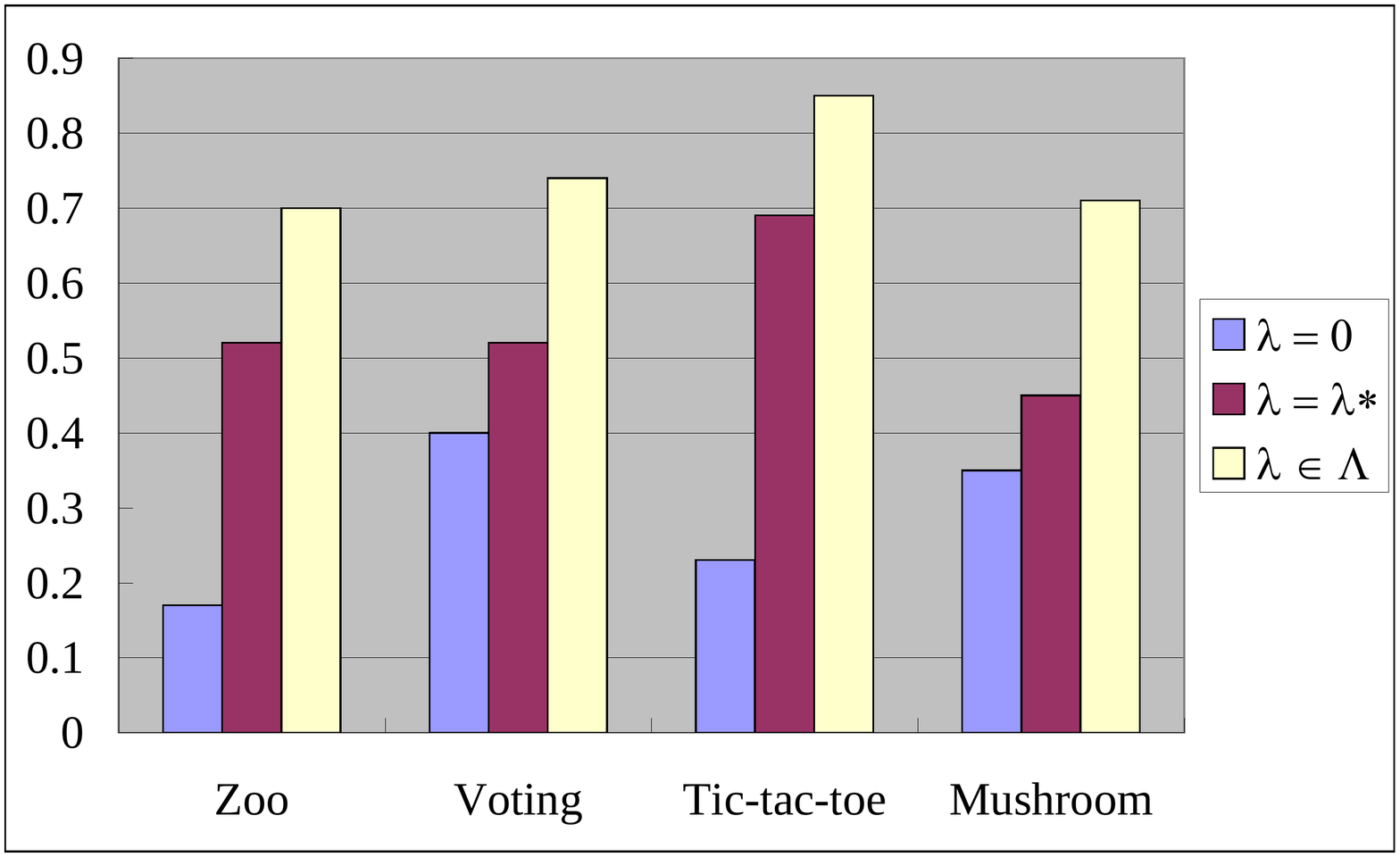}
    \caption{The probability of finding the optimal feature subset}
    \label{figure: heuristic}
    \end{center}
\end{figure}

General results are depicted in Fig. \ref{figure: heuristic}, from which we observe the following.
First, the approach without taking into considering the test cost performs poorly.
In most cases it cannot find the optimal feature subset.
Second, if we specify $\lambda$ appropriately, namely $\lambda = \lambda^*$, the results are more acceptable.
It is more likely to find the optimal feature subset.
However, as discussed earlier, we often have no idea how to specify it.
Third, the performance of the competition strategy is much better than the other two.
In more than 70\% cases it produces the optimal feature subset.
Moreover, the user does not have to know the optimal setting of $\lambda$.
In one word, the extra computation resource consumed by the competition strategy is worthwhile.

  %
  %%%%%%%%%%%%%%%%%%%%%%%%%%%%%%%%%%
  % beginning of a new section: experiments
  %%%%%%%%%%%%%%%%%%%%%%%%%%%%%%%%%%
  %
  \section{The CSP viewpoint to feature selection}\label{section: discussions}
Problems \ref{problem: reduct}, \ref{problem: minimal-reduct}, \ref{problem: minimal-test-cost-reduct} and \ref{problem: test-cost-constraint-reduct} provide the CSP viewpoint to feature selection.
Most existing feature selection problems in rough sets can be viewed extensions of Problem \ref{problem: minimal-reduct} in one or more of the following aspects: input, output, constraint, and optimization objective.
We analyze them from each aspect as follows.

First, there are some extensions concerning the input data model.
Since the data model is essential, these extensions often require extensions of the Pawlak rough set.
\begin{enumerate}
\item{Some conditional features are numeric.
Numeric data are quite different from symbolic ones which are employed in Pawlak rough sets \cite{Pawlak82Rough}.
Coverings, instead of partitions, can be formed according to feature sets.
Covering-based rough sets \cite{Zhu07Topological,ZhuWang03Reduction,ZhuWang07OnThree,YangLi10Reduction} deal with reduction of coverings.
The neighborhood rough set model \cite{HuQ2008MixedFeature,HuPedryczYuLang10Selecting,HuYuLiuWu08Neighborhood,HuYuXie08Numerical} generates neighborhood systems on such data.}
\item{The data are uncertain.
The uncertainty of data may be caused by noise, observational error, etc \cite{Dai12Uncertainty}.
The error range based covering rough set model \cite{MinZhu12Tcsdser} was proposed to deal with observational error.
Another well known data model might be interval-valued fuzzy sets \cite{GorzafczaryB1988Interval}, which has been studied through rough sets \cite{GongZ2008Rough}.}
\item{There are external information on features and feature subsets \cite{YaoZhaoWangHan08AModel}.
Some information are subjective and can be expressed by user preference.
For example, features are ranked by the user, or even directly specified by an expert \cite{MinLiuTanChen06TheMRelative}.
Other information are objective.
For example, there is a weight or test cost for each feature \cite{MinLiu09AHierarchical,YaoZhaoWangHan08AModel}.
There are a number of possible extensions to the weight computation of an feature subset.
These are additive, average, maximal, minimal extensions \cite{YaoZhaoWangHan08AModel}.
In \cite{MinLiu09AHierarchical}, six data models concerning test cost and relationships among features are defined.
Test-cost-sensitive attribute reduction problems \cite{HeMinZhu11Attribute,MinZhu12Tcsdser} can be defined according on these models.}
\item{There are external information on classification.
The most widely adopted information might be misclassification cost \cite{TurneyPD1995Cost,ZhouLiu06Training}.
DTRS \cite{LiYaoZhouHuang2009Two,YaoWong92ADecision,YaoZhaoWang08OnReduct} consider loss functions concerning different classifications.
These classifications correspond to positive, negative and boundary rules.
There are cost for both misclassifications and correct classifications.}
\item{There are external information on both conditions and classifications.
In applications such as clinic systems, both test costs and misclassification costs exist \cite{TurneyPD1995Cost}.
This issue is addressed in \cite{MinZhu12Tcsdser}.}
\end{enumerate}
%In fact, these data models have not been extensively studied.
%Other models should also be considered.

Second, there are some extensions concerning the output.
People considered generalized reduct problems, such as attribute value reduction \cite{Pawlak91Rough}, discretization \cite{Nguyen98Discretization}, symbolic value partition \cite{MinLiuFang08Rough}.
Since features are transformed or combined, these problems should be called \emph{feature extraction} instead \cite{GuyonI2006Feature,JainA1997Feature}.

Third, there are some extensions concerning the constraint.
Many of them are still expressed with the same form as Problem \ref{problem: minimal-reduct}.
However, the definitions of the positive region are different due to the change of the input data model.
Others are expressed with different forms.
\begin{enumerate}
\item{The computation of the positive region follows DTRS models \cite{YangYao12Modelling,YaoWong92ADecision,YaoZhaoWang08OnReduct}.
    In DTRS, parameters $\gamma$, $\beta$ and $\delta$ are used to define positive regions.
    They are in turn computed based on a set of loss functions according to the Bayesian decision procedure.
    The major advantage is that parameters are not set by the user subjectively.
    Therefore the models have good semantics and wide applications.}
\item{The computation of the positive region follows the variable precision rough set model \cite{Ziarko93Variable}, or the Bayesian rough set model \cite{SlezakZiarko06TheInvestigation}.
    There is a user-specified parameter $\beta$ to indicate the admissible classification error.
    Pawlak rough sets can be viewed a special case of variable precision rough sets where $\beta = 0$.
    This extension has inspired fruitful research works concerning probabilistic rough sets \cite{LiuLiRuan2011Probabilistic}.
    $\beta$-lower distribution and $\beta$-upper distribution \cite{MiWuZhang2004Approaches} have been more closely studied.}
\item{The computation of the positive region follows the neighborhood rough set model \cite{HuPedryczYuLang10Selecting,HuYuLiuWu08Neighborhood,HuYuXie08Numerical} or the error range based covering rough set model \cite{MinZhu12Tcsdser}.
    In the neighborhood rough set model \cite{HuPedryczYuLang10Selecting,HuYuLiuWu08Neighborhood,HuYuXie08Numerical}, positive regions also rely on a user specified parameter $\delta$, which is the distance threshold.
    In the error range based covering rough set model \cite{MinZhu12Tcsdser}, positive regions also rely on error ranges of data.
    Error ranges are determined by testing instruments and therefore they are objective.}
\item{The constraint is conditional information entropy \cite{Slezak02Approximate,Wang02Attribute,LiuLiMinYeYang05AnEfficient}.
    It is expressed by $H(B | \{d\}) = H(C | \{d\})$ where $H(B | \{d\})$ denotes the conditional information entropy of $B$ with respect to $d$.
    The conditional information entropy constraint is stricter than the positive region constraint.
    That is, the feature subset meeting the positive region constraint may not meet the conditional information entropy constraint.
    While the reverse does not hold.
    These two constraints are equivalent if and on if the decision system is consistent \cite{MiaoZhaoYaoLiXu09Relative}.}
\end{enumerate}

Fourth, there are some extensions concerning the optimization objective.
\begin{enumerate}
\item{Minimize the cost.
    In test cost sensitive decision systems, the objective is to minimize the total test cost \cite{MinHeQianZhu11Test}.
    In misclassification cost sensitive decision systems, the objective is to minimize the average misclassification cost \cite{LiHX11Further,YaoWong92ADecision,YaoZhaoWang08OnReduct}, or the risk \cite{LiZhou2011Risk,LiuLiRuan2011Probabilistic}.
    In decision systems with both test cost and misclassification cost, the objective is to minimize the total cost \cite{MinZhu11MinimalCost}.}
\item{Minimize the feature space $\prod_{a \in B}|V_a|$.
    For the minimal reduct problem, features with more values are more likely to be selected.
    These features, however, have weaker generalization ability than features with less values.
    The new objective can help amend this drawback.
    When the domains of features have the same size, the new objective coincides with Problem \ref{problem: minimal-reduct} \cite{MinDuQiuLiu07Minimal}.}
\item{Maximize the stability.
    Dynamic reducts \cite{BazanSkowron94Dynamic} are stable in the process of decision table sampling.
    Decision rules computed from dynamic reducts are more reliable.
    Parallel reducts \cite{Deng09Parallel} follow the same idea.}
\item{Maximize the margin.
    A margin is a geometric measure for evaluating the confidence of a classifier with respect to its decision \cite{CortesVapnik95Support,GiladR2004Margin}.
    Unlike other metric such as positive region or conditional information entropy, this measure is not monotonic.
    That is, it may increase or decrease when more features are selected.}
\end{enumerate}

Most problems mentioned above are no longer reduct problems.
When the input is changed, the indiscernibility relation may not exist.
One can only consider weaker relations such as the similarity relation \cite{SlowinskiVanderpooten00AGeneralized}.
When the constraint is changed, the positive region is not computed, or computed not in the Pawlak approach (see, e.g., \cite{HuYuLiuWu08Neighborhood,MinZhu12Tcsdser}).
Reducts subject to the conditional information entropy constraint may not be a Pawlak reduct.
When the optimization objective is changed, the optimal reduct may not be minimal.
Feature subset with the minimal total cost \cite{MinZhu12Tcsdser} may not be a reduct at all.

From these extensions, many meaningful new problems can be identified.
A few of them are listed as follows.
\begin{enumerate}
\item{Feature selection under DTRS with test cost.
    Note again the external information in DTRS cannot be expressed by a misclassification matrix.
    Test cost is also an external information.
    By considering more external information, the problem is more interesting and challenging.}
\item{Feature selection with positive region constraint.
    To have a even simpler representation, we may require the positive region to be preserved to a certain degree.
    For example, the feature subset should have a positive region more than 95\% of the original.
    Note that this problem is different from the variable precision rough set model \cite{Ziarko93Variable} where the definition of positive region is changed.
    Their motivations are, however, quite similar in that they all deal with the overfitting issue.}
\item{Minimal test cost feature selection with positive region constraint.
    This problem differs from the last one in that the objective is to find a feature subset with least cost.
    It is a hybrid of the last problem and the MTR problem.
    It can be also viewed a dual problem of the FSTC problem.}
\end{enumerate}

Some of these problems are new combinations of existing extensions, some involve new extensions.
We observe that the number of possible combinations is big, and many of them have certain application areas.
In other words, much research issues are opened from the CSP viewpoint.

  %
  %%%%%%%%%%%%%%%%%%%%%%%%%%%%%%%%%%
  % beginning of a new section: conclusions
  %%%%%%%%%%%%%%%%%%%%%%%%%%%%%%%%%%
  %
  \section{Conclusions and further works}\label{section: conclusion}
This paper proposed a new feature selection problem concerning the test cost constraint.
The new problem, called FSTC, has a wide application area because the resource one can afford is often limited.
Both backtracking and heuristic algorithms were designed for it.
Experimental results showed the efficiency of the backtracking algorithm compared with existing ones, and the effectiveness the competition strategy based on the $\lambda$-weighted heuristic algorithm.
It should be noted that with the competition strategy, we do not have to know the optimal setting of $\lambda$.
Instead, we can specify a set of $\lambda$ values which are valid for any dataset.

A more important contribution of the paper is the CSP viewpoint to feature selection in rough sets.
From this viewpoint, most feature selection problems are natural generalizations of the minimal reduct problem.
This viewpoint helps us to identify some other meaningful problems from the following aspects: input, output,
constraint, and optimization objective.
In summary, this paper has indicated important research trends concerning feature selection beyond rough sets.

  %
  %%%%%%%%%%%%%%%%%%%%%%%%%%%%%%%%%%
  % beginning of a new section: conclusions
  %%%%%%%%%%%%%%%%%%%%%%%%%%%%%%%%%%
  %
  \section*{Acknowledgements}\label{section: acknoledgements}
This work is supported in part by the National Natural Science Foundation of China under Grant No. 61170128,
the Natural Science Foundation of Fujian Province, China under Grant Nos. 2011J01374 and 2012J01294,
and the Science and Technology Key Project of Fujian Province, China under Grant No. 2012H0043.
  %
  %Reference
  %
%\bibliographystyle{../format/elsevier/elsart-num-sort}
%\bibliography{../bibitems/Roughsets,../bibitems/fanminRough,../bibitems/dtrs,../bibitems/nonRoughsets,../bibitems/cost}

\end{document}